# Adapting Legacy Robotic Machinery To Industry 4.0: A CIoT Experiment – version 1

Hadi Alasti, Purdue University Fort Wayne

## Abstract

This paper presents an experimental adaptation of a non-collaborative robot arm to collaborate with the environment, as one step towards adapting legacy robotic machinery to fit in industry 4.0 requirements. A cloud-based internet of things (CIoT) service is employed to connect, supervise and control a robotic arm's motion using the added wireless sensing devices to the environment. A programmable automation controller (PAC) unit, connected to the robot arm receives the most recent changes and updates the motion of the robot arm. The experimental results show that the proposed non-expensive service is tractable and adaptable to higher level for machine to machine collaboration. The proposed approach in this paper has industrial and educational applications. In the proposed approach, the CIoT technology is added as a technology interface between the sensors to the environment and the robotic arm. The proposed approach is versatile and fits to variety of applications to meet the flexible requirements of industry 4.0. The proposed approach has been implemented in an experiment using MECA 500 robot arm and AMAX 5580 programmable automation controller and ultrasonic proximity wireless sensor.

## Introduction

The market penetration of Industry 4.0 has increased in the past recent years, however many factories that have a large number of working legacy automation manufacturing equipment such as robot arms, assembly machines, etc. cannot afford whole system replacement. The financial problem caused by the slowed down economy has turned it to a bigger dilemma. Adapting the existing legacy programmable automation machinery to industry 4.0 by adding the required connectivity to industrial internet of things (IIoT), would be one non-expensive solution to the above requirement. Similar needs also have been reported as a push towards adapting the legacy systems in a transitional model (Rosendahl, 2015). Guidelines for adapting the legacy devices to industry 4.0 during the transition phase has been given in literature (Illa, 2018). This paper is presenting the experimental result of using cloud-based IoT between the programmable legacy automation machinery and the environment, that in a more general form can be extended to interaction between the manufacturing machines such as robots and the supervisory and control system.

Programmable logic controllers (PLC) have been the workhorse for the industrial automation for decades (Chivilikhin, 2020), however their legacy programming language is not necessarily adaptable to variety of robotic machines without a custom-designed hardware interface. Programmable automation controllers (PAC) with their versatility, computational power and networking options are one suitable replacement for PLCs in supervisory, control and networking of robotic machinery. Even though employing a PAC instead of PLC and regular PC provides much more flexibility, however collaboration with the environment can be confusing, as it does not provide a clear model to facilitate actual implementation of the legacy devices in industry 4.0 network. A model for transition of the legacy automation devices to industry 4.0 by using cloud-based internet of things (CIoT) was described by Delsing (Delsing, 2017). In a larger context, the development of interface between enterprise and control system has been discussed in ISA-95 (Williams, 1994).

It is worthy to mention that connecting the legacy devices using CIoT opens them to their cyber-physical shortcomings (Pessoa, 2018). Cloud security is just one of these issues, however solutions have been proposed for this shortcoming in literatures such as (Hamilton, 2017; Li, 2019; Sanislav, 2017; Tian, 2019). To assure security in wireless local area network (WLAN), in June 2018, WiFi Alliance and IEEE released WiFi protected access rev. 3 (WPA3) that allows customized security for IoT applications (Kwon, 2020).

In collection of environment and the machines' signals for collaboration, big data attributes such as variety, velocity, veracity, etc. are suitable to discuss. Accordingly, a big data gathering framework is required to be defined carefully, case by case for smart factory management (Bellavista, 2019). For instance, in collection of data from different sensors for quantities such as, distance, speed, dust, angle, touch, moisture, pressure, light intensity, gas density, etc. in a manufacturing site, besides their number, the speed and the quality of collection are also important. Accordingly, the problem of promoting a legacy robotic site can be an important big data problem (Xian, 2020; Martins, 2019).

An experimental integration of wireless sensing and CIoT to turn a non-collaborative robot arm to collaborate with the environment is explored, in this paper. The integration results show that employing a PAC and cloud-based service provide this capability to make the robot arm to collaborate with the environment. In broader view, this integration may allow the robot arms to collaborate to each other and to the environment, where it results in higher productivity, safer production environment, better products and less capital expenses. In this paper, an ultrasonic proximity sensor was used to monitor the distance of the approaching object in order to control the motion speed of the robot arm. The proximity sensor sends its observation of the distance to the running CIoT service. The PAC reads the new update of the sensor observation and controls the speed of the motion of the robot arm, in order to reduce the risk of incident with the approaching object. The discussed experiment in this paper has applications in industry and education. For instance, several companies use or produce products for cloud-based IoT, such as the system that MAN uses for tracking the engine faults in trucks and buses; smart factories automated by Siemens using industrial cloud based IoT; ABB considers sensors inside their robot for maintenance and repair of their products. For educational application, students in their curriculum become involved in robotics' concepts and also take courses in cloud computing and IoT. The proposed experience of this paper is a good merger between these three fields and make the students familiar with an area that is becoming a need in industry. Figure 1, illustrates the proposed approach in big picture.

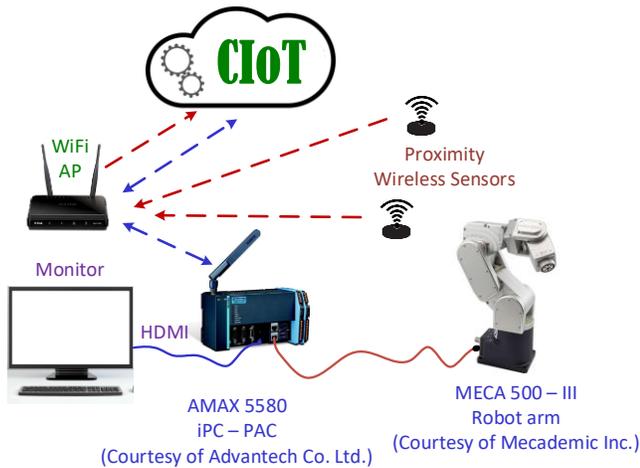

*Figure 1. The programmable automation controller (PAC) is attached to the MECA 500 robot arm to control its motion, based on the environment data that it reads from CIoT.*

The proposed approach in this paper has no conflict with open-source robot operating system (ROS) and eventually can be linked to extend the application domains of ROS. Similar to ROS (Guizzo, 2017) that allows the users to shift their software stacks to fit their robot and application area, here the user can add their own code for interface, protocol, etc.

The rest of this short paper is organized as follows. In the next section, the problem statement is given. Then, the proposed solution is discussed. After that, the implementation phase and also the performance of the CIoT-based data log is presented, in brief.

## Problem Statement

To adapt the legacy machinery of the robotic sites with industry 4.0 requirements, a pilot project was conducted. Wireless sensors such as ultrasonic proximity sensors shared their collected data from the environment to the robot control unit. Traditionally the control unit was a custom designed robot-controller that usually had insufficient capability to communicate with the environment or other brands of robots. The objective of this project was to add sensors to the environment to let a robot collaborate with the environment and specifically to control its motion's speed proportional to the distance of the approaching object, or completely stop the robot arm.

In the pilot project MECA 500-III robot arm (MECA 500) and AMAX 5580 (AMAX 5580) were used, which are non-collaborative mini robot arm and, an APC unit, respectively.

## The Proposed Solution

As mentioned, the MECA 500 robot arm is controlled using a PC. This robot arm is programmable either using a web interface that is accessible in a browser or independently using a high-level programming language such as Python. The robot arm has no sensor or camera to get information from the environment or any other working machines. In the proposed solution a modular PAC and sensor unit(s) were considered to allow the robot to receive real-time information from the environment.

In the proposed solution, the PC was replaced with one AMAX 5580. This unit is a non-expensive modular, industrial personal computer (iPC) with PAC feature. The iPC has several ports such as USB, RJ45, EtherCat, RS232 and RS485 ports, HDMI, etc. The modular feature of this new iPC allows to add additional ports, in order to control several devices, simultaneously.

In this project, a Python code was developed to program, control and monitor the robot arm in AMAX-5580 platform. The code controls the motion speed of the robot arm, and once an object approaches the robot arm, one or two of the ultrasonic proximity sensors may send the distance of the approaching object to the CIoT site. The iPC frequently and periodically checks for the most recent distance of the approaching object from the recorded sensor data over the CIoT and adapts the speed of the robot accordingly, in order to avoid or reduce the damage from incident.

In the discussed scenario in this paper, the sensors do not communicate with each other. The collected data from the sensors are stored on CIoT and analyzed or refined by the running code over the cloud. The continuously running code on PAC retrieves the refined data to use for environment awareness and controlling the speed of the robot arm.

The proposed approach improves work environment's safety by adding awareness to the robot arm control system. The robot control system changes the speed of the moving arm, or completely stops it, as a moving object (a human being, or another working robot) enters into a different neighborhood zones of the robot arm. The original motivation for this research is the tragic event that happened in Volkswagen factory in 2015, where a robot "in setup process" killed a worker (Washingtonpost, 2015).

In sensing side, one ultrasonic proximity sensor attached to one ESP32 microcontroller board, where the WiFi, Bluetooth and LoRaWAN connectivity allowed it to send the sensor observations to the remote CIoT. The application of LoRaWAN for industrial IoT has been already investigated (Sisinni, 2020). This wireless technology, which is unlicensed due to its low power transmissions at low data rates, is becoming more popular for industrial applications when the data volume is small. WiFi wireless connectivity is also an excellent option specifically for short range indoor applications. Due to the indoor location of the robot arm, in this project, WiFi was used for access to internet.

Based on the conducted measurements, the accuracy of the wireless sensors is around 1 cm and the wireless sensor assembly can measure the distance up to around 5 m. This accuracy and also the defense range are proper for reaction in medium to low motion speeds in indoor manufacturing sites.

In a general scenario, several wireless sensors at different angles will be placed around the robot arm. The simultaneous monitoring of the robot arm's surrounding also allows the iPC to detect the direction of approach of the moving object, which can be another moving robot in the manufacturing site. Figure 2, illustrates this scenario in development of a smart, invisible defense for the robot arm.

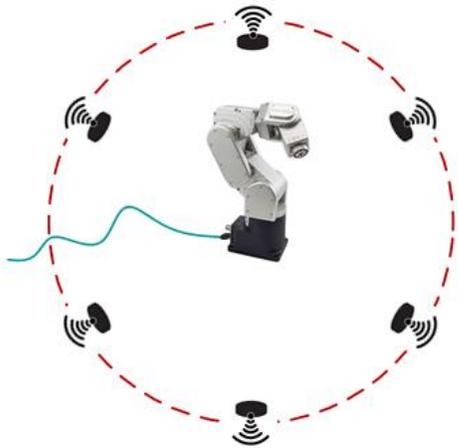

*Figure 2. Using several wireless sensors to make a smart, invisible fence around the robot arm.*

The general steps of the algorithm to the proposed solution is given in the following. According to this algorithm, the iPC continuously watches for any new updates in the data-log for any possible approaching objects in order to properly change the speed of the robot arm, based on the its distance and zones.

Design of the Python code on iPC is extremely important, as it needs to use interrupts or multi-tread programming to make sure that the robot arm reacts in time.

*Algorithm1: The implementation of the proposed algorithm*

1. Wireless proximity sensors sense the distance of the approaching object.
2. The presence of the approaching object is reported via WiFi to the running service over the cloud (CIoT).
3. The reported distance related to each sensor is stored in database over the cloud and presented in graph versus time.
4. The PAC collects each new update from the CIoT database.
5. The speed of the robot arm is controlled proportionally, if the objects motion and the robot arm are in risk of collision.
6. Repeat the process from step 1.

# Experimental Implementation

This project was implemented in four phases of i) searching for suitable sensors, and communication devices,0 and also approaches for implementation of environment sensing; ii) finding a cost efficient implementation for access to CIoT services; iii) purchasing an affordable PAC as a replacement for PLC and PC to work in industrial environment; iv) searching for a capable programming language that has potential networking capability to retrieve the stored data on CIoT database, and can be used to program the robot arm.

In the first phase of the project, several break-out microcontroller boards with wireless communication capabilities were considered as candidates, and finally ESP32 microcontroller due to its several wireless communication ports was selected. Also, among a large number of sensors, ultrasonic proximity sensor was selected due to its simplicity and low cost. For the first implementation phase of the project, only one wireless ultrasonic proximity sensor was attached to one ESP32 microcontroller board.

In the second phase of the project, the provisioning requirements of CIoT services were reviewed, such as leasing web space, creating online database, programming for online visualization of the data, and security for private access to the stored data. Against the development of the above items, there was another approach to use the available CIoT services. After preparing a list of the CIoT service providers, and considering factors such as simplicity of work with the service, cost, security, visualization, the allowed access speed and custom design capabilities based on common programming languages, Thingspeak was selected. For this project, one IoT channel was created on Thingspeak, which is easy to build, non-expensive and has flexible options. Thingspeak is affiliated with MathWorks and allows MATLAB modules and codes to be used for analysis of the collected data.

The third phase of the project was on selecting a versatile robot controller unit with access capability to internet and online databases. After consulting with a number of industries and vendors and getting no reasonable answer that met the available budget, an extensive search was conducted, where at the end the researches converged to purchasing one unit of AMAX 5580 from Advantech Co. Ltd. At the purchase time in August 2020, less than 100 units of this PAC were sold, worldwide.

The assembly of the robot arm, the AMAX 5580 PAC, and also the wireless proximity sensor module, are illustrated in Figure 3. The PAC module is connected to the robot arm via its RJ-45 jack and the CAT-5 Ethernet cable and works as master EtherCat module.

The PAC module is connected to internet via WiFi. The temporary assembly of the proximity wireless sensor node on breadboard is seen in the bottom of Figure 3. The wireless sensor sends its observations of the approaching object to one Thingspeak channel. Figure 4, illustrates the snapshot of some reported data on Thingspeak.

is comparable with preparing open-source personal code for ROS.

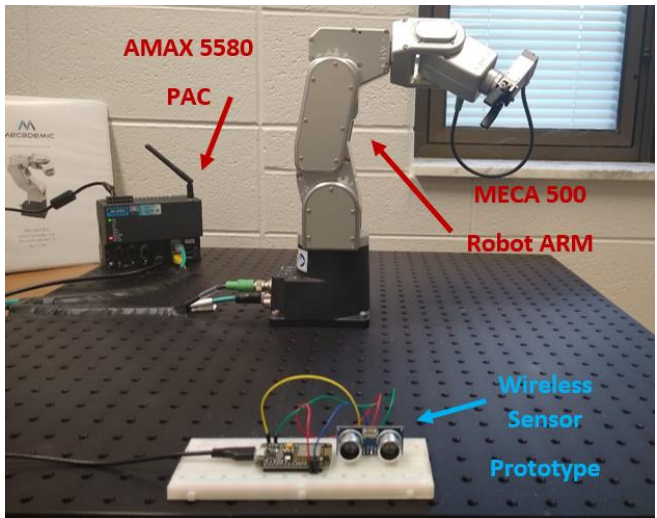

*Figure 3. The first implementation of the CIoT-based legacy robot control system.*

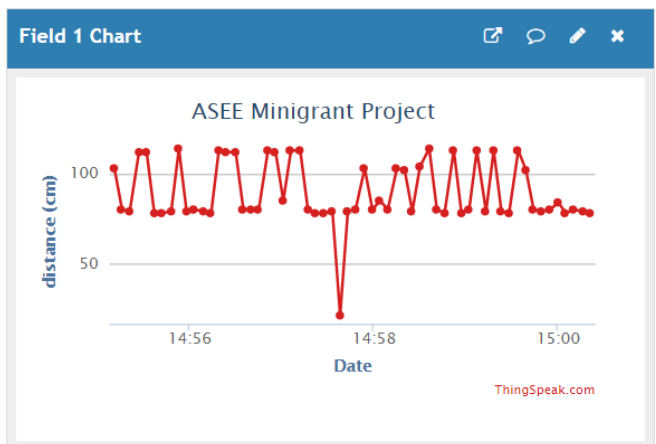

*Figure 4. The snapshot of the uploaded distance of the approaching object to the robot arm by one ultrasonic wireless sensor.*

In the final implementation phase of the project, several programming languages were considered, where among them, Python was selected due to its network programming capabilities, ease of programming and available modules.

The running Python code on iPC, periodically and frequently captures any new updates on the related channel of Thingspeak, and proportional to the distance to the robot arm to adjusts the speed of robot's motion. In a real-life example and in large scale, this control can avoid or at least reduce the risk of casualty in the work area.

Even though MECA 500 is not a legacy device and it has had limited installation for industry use, however it is worthy to mention that the proposed approach of this manuscript can be extended for the other forms of legacy robots, provided that two requirements are met. First, a connector interface between the legacy robot's connector to fit into the PAC input connector (if PAC does not have it). Second, a software interface is created that works as interface between the PAC and the robot. In the presented experiment of this article, Python language was used and the running Python code communicated with the robot arm via the EtherCat cable. The proposed approach in the presented experiment of this paper

## Conclusion and Future Works

An experimental implementation to use Cloud-based Internet of Things (CIoT) as technology interface for adapting the legacy automation devices to Industry 4.0 is investigated and discussed. MECA 500, as a non-collaborative robot arm is adapted to collaborate with the environment by adding AMAX-5580, which is a programmable automation controller (PAC) as programming and control unit, and one wireless sensor for environment observation. In an experimental implementation, one ultrasonic proximity sensor was added to monitor the surroundings of the robot arm to collect the distance information about the approaching objects. The wireless sensor data was recorded on one channel of Thingspeak, as CIoT service provider. A Python code which ran on PAC collected the recorded information of the wireless sensors in real time, in order to adjust the speed of motion of the robot arm. In larger scale, this project is implementable for collaboration between two or more robots or machines. According to this experiment, it is possible to adapt the legacy robotic devices to industry 4.0 by adding wireless (or wired) sensors to the system or environment and using the collected information on CIoT to control the robots and machines. The proposed approach has industrial as well as educational applications.

In the next step of this project and as future work, the use of network of things (NoT) instead of CIoT is under investigation. Also, to improve the decisioning system, using machine learning algorithms in merger of the collected sensor's data will be considered. Another possible focus area of this research is the merger of the proposed approach with open-source ROS. Latency in communication and decisioning is another aspect that can be evaluated by the use of 5G and next generation networks.

## Acknowledgement

The author would like to appreciate Purdue University Fort Wayne, School of Polytechnic, and American Society of Engineering Education (ASEE) for their financial supports. This paper has been partly supported by ASEE ETD Minigrant.

The author would like to appreciate the anonymous reviewers for their constructive comments and their valuable time.

# Biographies

**HADI ALSTI** is an assistant professor at School of Polytechnic of Purdue University Fort Wayne. He earned his PhD in electrical engineering from the University of North Carolina in 2009. Dr. Alasti's research interests include wireless communications, wireless sensor networks, emerging communication technologies and energy efficient approaches. Dr. Alasti also has worked for several years in power system and power system communications and has published more than 30 journal articles and conference papers. Dr. Alasti may be reached at: alastih@pfw.edu